\documentclass[11pt]{article}
\usepackage{authblk}
\usepackage{graphicx} 
\usepackage[margin=1in]{geometry}  
\usepackage[hidelinks]{hyperref}
\usepackage[labelfont=bf]{caption} 
\usepackage{nameref}
\usepackage{booktabs}  
\usepackage{longtable} 
\usepackage{array}     
\usepackage{float}
\usepackage{makecell}
\usepackage{multirow}
\usepackage{titlesec}
\usepackage{wrapfig}

\titleformat{\subsection}[hang]{\bfseries\normalsize}{\thesubsection}{1em}{}
\titleformat{\subsubsection}[hang]{\itshape\normalsize}{\thesubsubsection}{1em}{}

\title{AI-Driven Detection and Analysis of Handwriting on Seized Elephant Ivory}

\author[1]{Will Fein}
\author[2]{Ryan J. Horwitz}
\author[2]{John E. Brown III}
\author[1]{Amit Misra}
\author[1]{Felipe Oviedo}
\author[1]{Kevin White}
\author[1]{Juan M. Lavista Ferres}
\author[2]{Samuel K. Wasser}

\affil[1]{AI for Good Lab, Microsoft}
\affil[2]{Center for Environmental Forensic Science, University of Washington}

\date{}  

\begin{document}

\maketitle

\begin{abstract}

The transnational ivory trade is driving the decline of elephant populations across Africa, but trafficking networks remain difficult to disrupt. Tusks seized by law enforcement officials carry forensic information on implicated traffickers. This includes handwritten markings, which are easy to photograph, but are rarely documented or analyzed. Here, we present an AI-driven pipeline for extracting and analyzing handwritten markings on seized elephant tusks to offer a novel, scalable, and low-cost source of forensic evidence. Our dataset contains 6,085 photographs from eight large seizures of ivory made over a 6-year period (2014-2019). We used an object detection model to extract over 17,000 individual markings, which were then labeled and described with state-of-the-art AI tools. From this evidence, we established vital new connections across ivory shipments. Our approach demonstrates the transformative potential of AI in wildlife forensics and highlights practical steps for integrating handwriting analysis into efforts to disrupt organized wildlife crime.

\end{abstract}

\section*{Introduction}

Each year, poachers kill tens thousands of African elephants for their tusks in hotspots across Africa. Middlemen then move the ivory in small quantities across national borders where they are consolidated into large shipments for international export by transnational criminal organizations (TCOs) \cite{UNODC2016}. 

Wasser et al. pioneered methods that use DNA evidence from elephant tusks in seized shipments to determine where elephants were poached \cite{Wasser2007, Wasser2008, Wasser2009, Wasser2015} as well as the interconnected nature of the TCOs driving the trade \cite{Wasser2018, Wasser2022}. However, the high cost of DNA analysis has limited investigations to a representative sample of tusks in multi-ton seizures (10-50\% of the tusks), potentially missing crucial information that could lead to new connections or support existing ones. 

Signatures handwritten on tusks can fill these gaps. The potential value of this data source is well-documented; it is believed that these signatures are made by middlemen to assure that they are paid for their efforts once a shipment is sold \cite{UNODC2014}. Handwriting on tusks can be readily photographed, providing an inexpensive source of forensic data. However, until now, photographs of tusk markings have been underutilized due to the lack of a scalable approach. Computer vision and AI tools can unlock the value of handwriting data.

Between 2014 and 2019, the University of Washington’s Center for Environmental Forensic Science (CEFS) collected 6,085 photographs of tusks from 8 seized shipments, each containing at least 1,100kg of unworked ivory. In this study, we fine-tuned an AI vision-language model to detect handwriting in photographs of seized tusks and implemented a semi-automated pipeline that applies meaningful labels to those handwritten markings. We linked shipments by matching “signature markings” that appeared on one or more tusks in multiple seizures. This project introduces an inexpensive, scalable method for extracting forensic evidence from tusk markings that expands the ability of law enforcement to uncover connections among ivory seizures and trafficking networks.

\begin{figure}[!t]
  \centering
  \fbox{\includegraphics[width=0.9\textwidth]{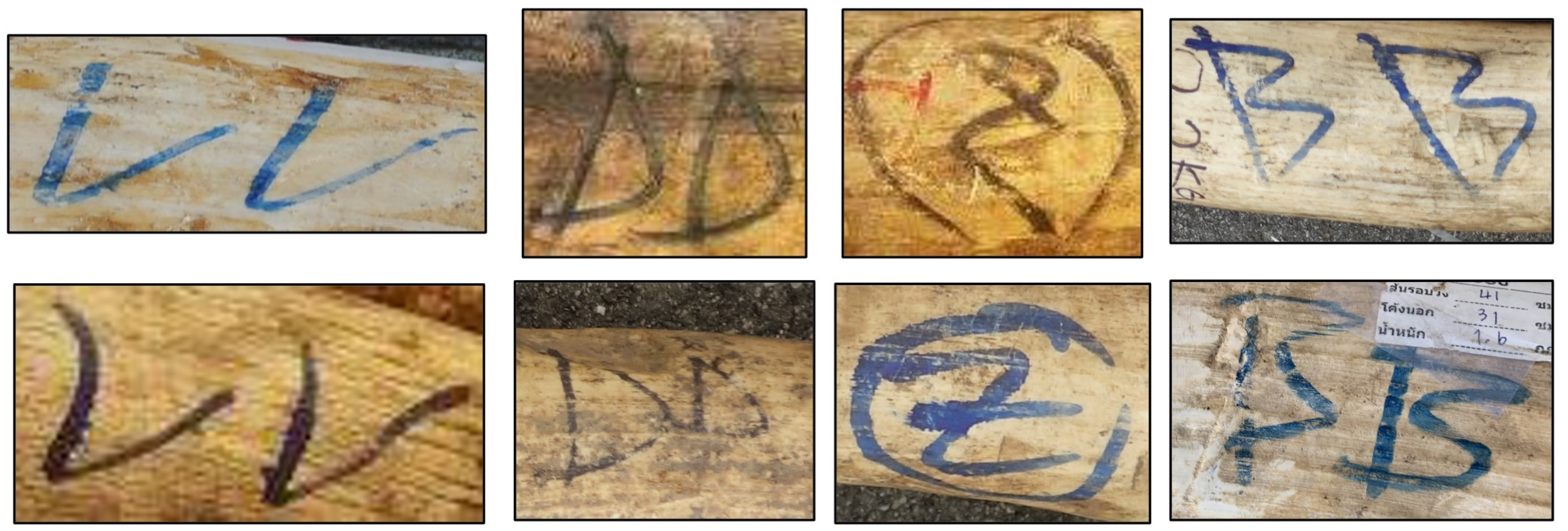}}
    \caption{Examples of matched signature markings. Each column shows markings found in two different seizures.}
  \label{fig:Fig1}
\end{figure}

\section*{Results}

From 6,085 photos, we extracted and labeled 17,071 handwritten markings with a novel semi-automated pipeline, released as an open-source repository. 44\% of markings were determined to be made post-seizure by government authorities (see Materials and Methods), leaving 9,575 pre-seizure markings presumed to be written by traffickers.
 
We defined “signature markings” as those with sufficient consistency and specificity to connect the tusks on which they appear. We identified 4,860 unique signature markings; 184 of the 4,860 were observed more than once, suggesting that many different individuals handled the tusks as they moved up the supply chain. Twenty of the 4,860 signatures appeared in multiple seizures, representing novel evidence of connections among ivory shipments.

We identified 1,168 markings containing two letter initials. Sixteen occurred at least 10 times, accounting for 75\% of the observed initials markings. The two most common initials were observed 239 times and 169 times respectively. The small number of initials occurring with high frequency suggests that a small number of high-level middlemen are consolidating tusks from many lower-level middlemen prior to delivery to the exporters.

\begin{figure}[ht]
  \centering
  \fbox{\includegraphics[width=0.9\textwidth]{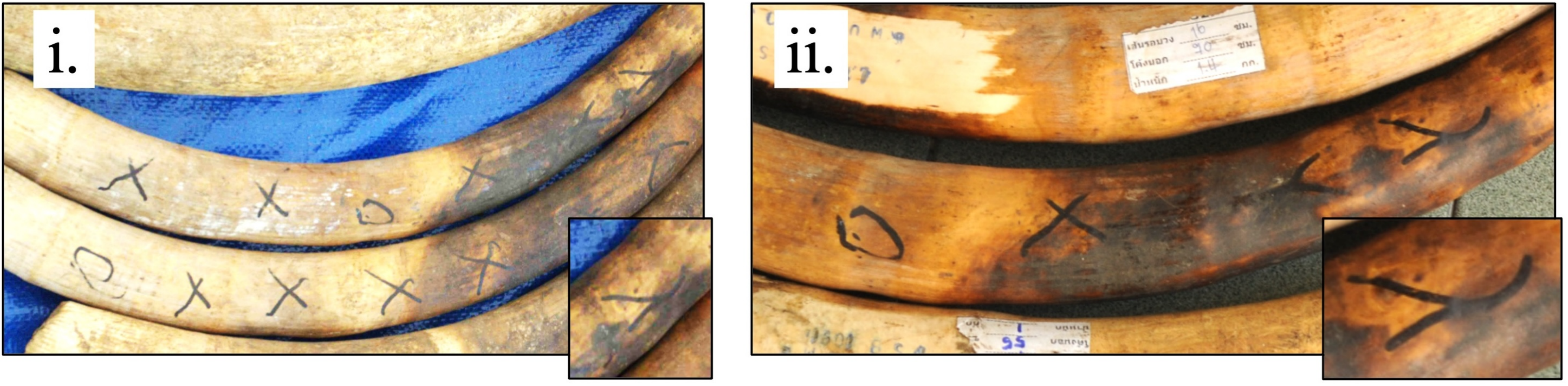}}
    \caption{Examples of the lambda symbol. Image i. is from seizure 2 and ii. is from seizure 3.}
  \label{fig:Fig2}
\end{figure}

We also investigated other markings of interest, investigations which were made possible by the descriptive labels created by our pipeline. Sequences of X’s and O’s appear in seizures 2, 3, 4, 5, 7, and 8, but their ubiquity and heterogeneity prevented the automatic identification of a strong evidentiary connection among all six seizures. Ivory traffickers used these markings regularly, but distinct entities may have been responsible for different instances of the pattern. Manual examination by domain experts revealed a distinctive variant: sequences that end with a lambda symbol (Figure \ref{fig:Fig2}). Found in seizures 2 and 3, these establish a seventh cross-seizure connection.

Notably, four of the seven connections established through handwriting evidence involve seizure 2, a seizure without available genetic data.

\begin{table}[t!]
\centering
\caption{Cross-Seizure Connections Established with Handwriting Evidence}
\begin{tabular}{ll}
Seizure Grouping & Handwriting Evidence \\
\midrule
Seizures 2 \& 3 & Lambda X/O Variation \\
Seizures 2 \& 8 & 7 shared signature markings \\
Seizures 3 \& 8 & 1 shared signature marking \\
Seizures 5 \& 8 & 8 shared signature markings \\
Seizures 7 \& 8 & 1 shared signature marking \\
Seizures 2, 5 \& 8 & 1 shared signature marking \\
Seizures 2, 7 \& 8 & 1 shared signature marking \\
\bottomrule
\end{tabular}
\label{tab:Table1}
\end{table}

\section*{Discussion}
This work demonstrates how AI and computer vision tools can unlock the evidentiary potential of handwritten signatures on tusks in the illicit elephant ivory trade. Photographing these signatures is cost-effective and simple, only requiring a camera or smartphone. No other method for collecting forensic evidence related to the transnational ivory trade is as straightforward and scalable. 

DNA analyses establish the strongest possible links between seizures by matching tusks from the same elephant or close family members \cite{Wasser2022}. However, financial, legal, and temporal constraints may limit the number of tusks that can be genetically analyzed. Handwriting analysis can fill these gaps. While seizure 2 was never genotyped, handwriting analysis identified 10 shared signature markings between seizures 2 and 8, suggesting a connection.

Moreover, while genetic samples are collected from representative samples of tusks, photographs can cost-effectively capture every tusk in a seizure. This is critical, as handwriting-based connections may rely on finding one marking among thousands. The “Circled Z” marking (Figure \ref{fig:Fig1}) occurred 151 times in seizure 5 but was identified only once in seizure 8. DNA matching of those two seizures failed to detect that link \cite{wasser2025ivory}. Combining handwriting evidence with DNA data and other forms of physical evidence can illuminate previously hidden connections and provide deep insights into the transnational ivory trade \cite{wasser2025ivory}. 

Although our initial dataset exclusively contained photographs from large ivory seizures (at least 1.1 tonnes), photographs of illicit ivory can be taken in many contexts, including smaller collections of tusks seized between the source and export locations. Performing handwriting analysis at multiple points along the supply chain could help identify transport routes used by ivory traffickers and determine what stage along the supply chain the signatures – and corresponding criminal actors – first appear.

This study has limitations. While our method enables tracking of signature markings across multiple tusks, it cannot currently confirm whether any two signature markings were made by the same individual. It is unlikely that even an expert human reviewer could confidently link individuals to multiple texts \cite{Hicklin2022}, let alone signatures with so few characters and no reference text. Additionally, the lack of textual context, textured backgrounds, and low-quality handwriting in our images pose substantial challenges for large language models performing character recognition. That said, GPT-4o was able to provide sufficiently high-quality output to enable powerful conclusions. This pipeline will become more effective as LLMs continue to improve character recognition tasks.

The illegal ivory trade has become highly transnational and shows little sign of stopping. Handwriting analysis is a powerful tool that can fill gaps in the information provided by other forensic techniques. It can rapidly and inexpensively generate vast amounts of data to improve our understanding of how ivory moves up the supply chain. Hopefully, this paper will encourage others to help expand this database.

\section*{Materials and Methods}
\label{sec:methods}

The pipeline has four components. First, individual markings are extracted from photographs using a computer vision model. Second, human reviewers label a subset of markings from each seizure. Third, the new labels are automatically applied to similar markings using traditional machine learning techniques. Fourth, a large language model (LLM) labels the remaindering markings and provides natural language descriptions of all markings. These descriptions provide detailed information about each marking, including exact text and descriptions of handwriting style. 

We performed handwriting detection using a version of the MM-GROUNDING-DINO vision/ language model fine-tuned on a dataset of 374 manually annotated images from our dataset \cite{Zhao2024, Liu2023}. We conducted model evaluation using a randomly selected test set of 94 images not exposed to the model during training. We selected the model version with the best out-of-sample recall because our principal goal was to avoid missing any markings. The model achieved 96\% recall and 94\% precision. We implemented a post-processing script to ensure consistent counting of distinct markings. Examples of model detections are in Figure 3. See the SI for more details.

After each sample of extracted markings was manually labeled, the labels were automatically applied to highly similar markings. The markings were first encoded into vectors using OPEN-CLIP \cite{Radford2021} and a support vector machine classifier was trained for each label class. Unlabeled images were assigned to a given class when the fitted probability of belonging to that class was at least 90\%. This phase of the pipeline was most successful in identifying markings made post-seizure, for example by authorities inventorying tusks as required by the Convention on International Trade in Endangered Species (CITES); these were more visually consistent than pre-seizure markings. In this step, 95\% of post-seizure markings were identified by an SVM. Only 38\% of the other markings were given a class label. Precision was calculated by manually assessing the correctness of a random sample of 400 markings. Labels assigned as post-seizure had a precision of 93\%. Other labels had a precision of 88\%.

\begin{figure}[t!]
\centering
\includegraphics[width=.6\linewidth]{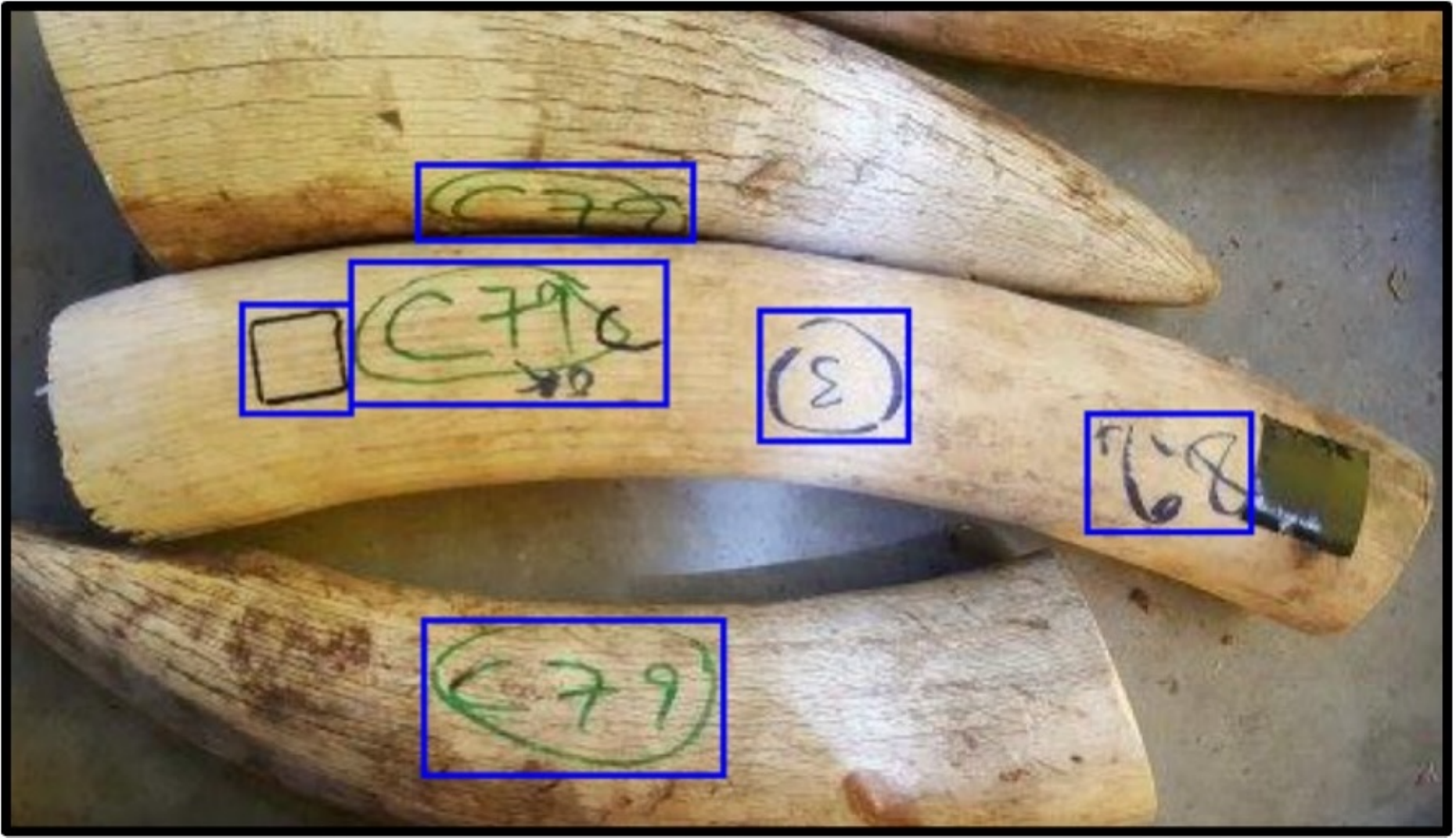}
\caption{Example of marking detections from the object detection model.}
\label{fig:Det}
\end{figure}

The remaining unlabeled markings were labeled using OpenAI’s GPT-4o large language model. The LLM was also used to provide additional information on markings that were already labeled, including exact text and natural language descriptions. More information, including exact prompts, can be found in the SI and the open-source repository: \url{https://github.com/microsoft/signature-marking-detector}.

The optical character recognition piece of this pipeline proved least accurate. State-of-the-art LLMs exhibit poor performance in character recognition when there is little contextual information or the text is handwritten \cite{ocrbench2023}. Our tasks included both challenges, and our text was sometimes cutoff, very faint, or nearly illegible. Despite these challenges, the overall character error rate (CER) is 0.265, as measured in a test set of 200 randomly selected markings. CER is 0.295 for single-character markings, which have no contextual information, and 0.260 for longer texts.

This level of performance was sufficient  to draw strong conclusions from handwriting data. The pipeline identified a set of matched signature markings confirmed through visual inspection, requiring only a fraction of the time that would have been required to examine and compare all 17,071 markings. As LLMs continue to improve, we expect the required human review time to decrease even further.

No human subjects or live animal research was conducted in this study.

\subsection*{Data and Code Availability}

The dataset of imagery is not publicly available as the photographs constitute law enforcement sensitive data. Model weights of the fine-tuned detection model are available at \url{https://zenodo.org/records/16423661}. Code for post-processing of the detections and semi-automated labeling, including all of the exact LLM prompts, is available at \url{https://github.com/microsoft/signature-marking-detector}.

\section*{Acknowledgments}
We thank Sadie Kahan for her assistance in data labeling during her internship at Microsoft.
\newpage
\bibliography{references}

\begin{thebibliography}{10}

\bibitem{UNODC2016}
{United Nations Office on Drugs and Crime}.
\newblock World wildlife crime report 2016.
\newblock Technical report, United Nations Office on Drugs and Crime, 2016.
\newblock Accessed August 11, 2025.

\bibitem{Wasser2007}
S.~K. Wasser et~al.
\newblock Using dna to track the origin of the largest ivory seizure since the 1989 trade ban.
\newblock {\em Proc. Natl. Acad. Sci. U.S.A.}, 104:4228--4233, 2007.

\bibitem{Wasser2008}
S.~K. Wasser et~al.
\newblock Combating the illegal trade in african elephant ivory with dna forensics.
\newblock {\em Conserv. Biol.}, 22:1065--1071, 2008.

\bibitem{Wasser2009}
S.~K. Wasser, B.~Clark, and C.~Laurie.
\newblock The ivory trail.
\newblock {\em Sci. Am.}, 301:68--76, 2009.

\bibitem{Wasser2015}
S.~K. Wasser et~al.
\newblock Genetic assignment of large seizures of elephant ivory reveals africa’s major poaching hotspots.
\newblock {\em Science}, 349:84--87, 2015.

\bibitem{Wasser2018}
S.~K. Wasser et~al.
\newblock Combating transnational organized crime by linking multiple large ivory seizures to the same dealer.
\newblock {\em Sci. Adv.}, 4:eaat0625, 2018.

\bibitem{Wasser2022}
S.~K. Wasser et~al.
\newblock Elephant genotypes reveal the size and connectivity of transnational ivory traffickers.
\newblock {\em Nat. Hum. Behav.}, 6:371--382, 2022.

\bibitem{UNODC2014}
{United Nations Office on Drugs and Crime}.
\newblock {\em Guidelines on Methods and Procedures for Ivory Sampling and Laboratory Procedures}.
\newblock United Nations publication, Laboratory and Scientific Section, 2014.

\bibitem{wasser2025ivory}
S.~Wasser et~al.
\newblock A higher-level network has supplied african elephant ivory to a revolving door of criminal networks for decades.
\newblock Manuscript submitted for publication to Proc. Natl. Acad. Sci., August 2025, 2025.

\bibitem{Hicklin2022}
R.~A. Hicklin et~al.
\newblock Accuracy and reliability of forensic handwriting comparisons.
\newblock {\em Proc. Natl. Acad. Sci. U.S.A.}, 119:e2119944119, 2022.

\bibitem{Zhao2024}
X.~Zhao et~al.
\newblock An open and comprehensive pipeline for unified object grounding and detection.
\newblock arXiv preprint, 2024.

\bibitem{Liu2023}
S.~Liu et~al.
\newblock Grounding dino: Marrying dino with grounded pre-training for open-set object detection.
\newblock arXiv preprint, 2023.

\bibitem{Radford2021}
A.~Radford et~al.
\newblock Learning transferable visual models from natural language supervision.
\newblock In {\em Proc. 38th Int. Conf. Mach. Learn.}, volume 139, pages 8748--8763, 2021.

\bibitem{ocrbench2023}
Y.~Liu et~al.
\newblock Ocrbench: On the hidden mystery of ocr in large multimodal models.
\newblock \textit{CoRR}, 2023.

\end{thebibliography}

\end{document}